\def\year{2022}\relax
\documentclass[letterpaper]{article} 
\usepackage{aaai22}  
\usepackage{times}  
\usepackage{helvet}  
\usepackage{courier}  
\usepackage[hyphens]{url}  
\usepackage{graphicx} 
\usepackage{natbib}  
\usepackage{caption} 
\DeclareCaptionStyle{ruled}{labelfont=normalfont,labelsep=colon,strut=off} 
\usepackage{algorithm}
\usepackage{algorithmic}

\usepackage{newfloat}
\usepackage{listings}
\usepackage{amsmath}
\usepackage{amssymb}
\usepackage{booktabs}
\usepackage{multirow}
\usepackage{makecell}
\floatstyle{ruled}
\newfloat{listing}{tb}{lst}{}
\floatname{listing}{Listing}
\title{End-to-End Transformer Based Model for Image Captioning}
\author{
    Yiyu Wang\textsuperscript{\rm 1},
    Jungang Xu\textsuperscript{\rm 2}\thanks{Corresponding Author},
    Yingfei Sun\textsuperscript{\rm 1}
}
\affiliations{
    \textsuperscript{\rm 1}School of Electronic, Electrical and Communication Engineering, University of Chinese Academy of Sciences\\
    \textsuperscript{\rm 2}School of Computer Science and Technology, University of Chinese Academy of Sciences\\

    wangyiyu18@mails.ucas.ac.cn, xujg@ucas.ac.cn, yfsun@ucas.ac.cn
}

\begin{document}

\maketitle

\begin{abstract}
CNN-LSTM based architectures have played an important role in image captioning, but limited by the training efficiency and expression ability, researchers began to explore the CNN-Transformer based models and achieved great success. Meanwhile, almost all recent works adopt Faster R-CNN as the backbone encoder to extract region-level features from given images. However, Faster R-CNN needs a pre-training on an additional dataset, which divides the image captioning task into two stages and limits its potential applications. In this paper, we build a pure Transformer-based model, which integrates image captioning into one stage and realizes end-to-end training. Firstly, we adopt SwinTransformer to replace Faster R-CNN as the backbone encoder to extract grid-level features from given images; Then, referring to Transformer, we build a refining encoder and a decoder. The refining encoder refines the grid features by capturing the intra-relationship between them, and the decoder decodes the refined features into captions word by word. Furthermore, in order to increase the interaction between multi-modal (vision and language) features to enhance the modeling capability, we calculate the mean pooling of grid features as the global feature, then introduce it into refining encoder to refine with grid features together, and add a pre-fusion process of refined global feature and generated words in decoder. To validate the effectiveness of our proposed model, we conduct experiments on MSCOCO dataset. The experimental results compared to existing published works demonstrate that our model achieves new state-of-the-art performances of 138.2\% (single model) and 141.0\% (ensemble of 4 models) CIDEr scores on `Karpathy' offline test split and 136.0\% (c5) and 138.3\% (c40) CIDEr scores on the official online test server. 
Trained models and source code will be released.
\end{abstract}

\section{Introduction}
Image captioning aims to automatically describe the visual content of a given image with fluent and credible sentences. It is a typical multi-modal learning task, which connects Computer Vision (CV) and Natural Language Processing (NLP). Inspired by the success of deep learning methods in machine translation \cite{DBLP:conf/acl/PapineniRWZ02, DBLP:conf/emnlp/ChoMGBBSB14}, almost all image captioning models adopt the encoder-decoder framework with the visual attention mechanism. The encoder encodes input images into fix-length vector features, and the decoder decodes image features into descriptions word by word \cite{DBLP:conf/cvpr/VinyalsTBE15, DBLP:conf/icml/XuBKCCSZB15, DBLP:conf/cvpr/00010BT0GZ18, huang2019attention, DBLP:conf/cvpr/PanYLM20}. 

Initially, researchers adopted a pre-trained Convolutional Neural Network (CNN) as an encoder to extract image grid-level features and Recurrent Neural Network (RNN) as a decoder \cite{DBLP:conf/cvpr/VinyalsTBE15, DBLP:conf/icml/XuBKCCSZB15}. 
\cite{DBLP:conf/cvpr/00010BT0GZ18} first adopted Faster R-CNN to extract region-level features. Due to its overwhelming advantage, most subsequent works followed this pattern, and grid-level features extracted by CNN were discarded. 
Nevertheless, there are still some inherent defects in region-level features and encoder of object detector: 1) region-level features may not cover the entire image, which results in the lack of fine-grained information \cite{DBLP:conf/aaai/LuoJSCWHLJ21}; 2) extracting region features is high time consuming, and the object detector needs an extra Visual Genome \cite{DBLP:journals/ijcv/KrishnaZGJHKCKL17} dataset for pre-training, which makes it difficult to train image captioning model end-to-end from image pixels to descriptions, and also limits potential applications in the actual scene \cite{DBLP:conf/cvpr/JiangMRLC20}.

Decoder of LSTM \cite{DBLP:journals/neco/HochreiterS97} with soft attention \cite{DBLP:conf/icml/XuBKCCSZB15} mechanism has remained the common and dominant approach in the past few years. However, the shortcomings of training efficiency and expression ability of LSTM also limit the effect of relevant models. Inspired by the success of Multi-head Self-Attention (MSA) mechanism and Transformer architecture in NLP tasks \cite{DBLP:conf/nips/VaswaniSPUJGKP17}, many researchers began to introduce MSA into decoder of LSTM \cite{huang2019attention,DBLP:conf/cvpr/PanYLM20} or directly adopt Transformer architecture as decoder \cite{DBLP:conf/cvpr/CorniaSBC20,DBLP:conf/cvpr/PanYLM20,DBLP:conf/aaai/LuoJSCWHLJ21,DBLP:conf/aaai/JiLSCLW0J21} of image captioning models.

Especially, Transformer architecture gradually shows extraordinary potential in CV tasks \cite{DBLP:conf/iclr/DosovitskiyB0WZ21,liu2021Swin} and multi-modal tasks \cite{DBLP:conf/nips/LuBPL19, DBLP:conf/cvpr/ZhuY20a,DBLP:conf/icml/RadfordKHRGASAM21}, which provides a new choice for encoding images into vector features. Different from Faster R-CNN, features extracted by a visual transformer are grid-level features, which have a higher computing efficiency and allows expediently exploring more effective and complex designs for image captioning.

Considering the disadvantage of pre-trained CNN and object detector in encoder and limitations of LSTM in decoder, we build a pure Transformer-based image captioning model (PureT) to integrate this task into one stage without pre-training process of object detection to achieve end-to-end training. In Encoder, we adopt Swin-Transformer \cite{liu2021Swin} to extract grid features from given images as the initial vector features and compute the average pooling of gird features as the initial image global feature. Then, we construct a refining encoder similar to \cite{huang2019attention,DBLP:conf/cvpr/CorniaSBC20,DBLP:conf/aaai/JiLSCLW0J21} by Shifted Window MSA (SW-MSA) from Swin-Transformer to refine image initial grid features and global feature. The refining encoder has a similar architecture with Transformer Encoder in machine translation \cite{DBLP:conf/nips/VaswaniSPUJGKP17} which can be regarded as an extension of Encoder of SwinTransformer for image captioning model. In Decoder, we directly adopt Transformer Decoder in machine translation \cite{DBLP:conf/nips/VaswaniSPUJGKP17} to generate captions. Furthermore, we pre-fuse the word embedding vector with the image global feature from Encoder before the MSA of word embedding vector to increase the interaction of inter-model (image-to-words) features.

We validate our model on MSCOCO \cite{DBLP:conf/eccv/LinMBHPRDZ14} offline ``Karpathy'' \cite{DBLP:journals/pami/KarpathyF17} test split and official online test server. The results demonstrate that our PureT achieves new state-of-the-art performance on both single model and ensemble of 4 models configurations: on offline ``Karpathy'' test split, a single model and an ensemble of 4 models achieve 138.2\% and 140.8\% CIDEr scores respectively; on official online test server, an ensemble of 4 models achieves 135.3\% (c5) and 138.0\% (c40) CIDEr. 

Our main contributions are summarized as follows:
\begin{itemize}
\item We construct a pure Transformer-based (PureT) model for image captioning, which integrates this task into one stage again without the pre-training process of object detector and provide a new simple and solid baseline of image captioning.
\item We add a pre-fusion process between the generated word embeddings and image global feature, which aims to increase the interaction of inter-modal features and enhance the reasoning ability from image to captions.
\item We conduct extensive experiments on the MSCOCO dataset, which demonstrate the effectiveness of our proposed model, and achieve a new state-of-the-art performance on both `Karpathy' offline test split and official online test server.
\end{itemize}

\section{Related Work}
Existing works of image captioning can be divided into CNN-LSTM based models \cite{DBLP:conf/cvpr/VinyalsTBE15,DBLP:conf/icml/XuBKCCSZB15,DBLP:conf/cvpr/00010BT0GZ18,DBLP:conf/aaai/WangC019,huang2019attention} and CNN-Transformer based models \cite{DBLP:conf/nips/HerdadeKBS19,GuangLi2019,DBLP:conf/cvpr/PanYLM20,DBLP:conf/cvpr/CorniaSBC20,DBLP:conf/aaai/JiLSCLW0J21,DBLP:conf/aaai/LuoJSCWHLJ21}. Both adopted pre-trained CNN or Faster R-CNN as the encoder to encode image into grid or region-level features, where the former models adopted Long Short-Term Memory Network (LSTM) \cite{DBLP:journals/neco/HochreiterS97} as the decoder and the latter models adopted Transformer \cite{DBLP:conf/nips/VaswaniSPUJGKP17} as the decoder to generate description word by word.

Earlier works used pre-trained CNN, e.g., VGG-16 \cite{DBLP:journals/corr/SimonyanZ14a} and ResNet-101 \cite{DBLP:conf/cvpr/HeZRS16}, as the encoder to encode image into grid-level features with fixed-length, and then LSTM with attention mechanism was applied among them to generate captions \cite{DBLP:conf/icml/XuBKCCSZB15,DBLP:conf/cvpr/RennieMMRG17}. \cite{DBLP:conf/cvpr/00010BT0GZ18} first introduced Faster R-CNN \cite{DBLP:conf/nips/RenHGS15} into image captioning to extract the region-level features more in line with the human visual habits, which has become a typical pattern to extract image features in subsequent works.

Above all models adopted LSTM as the decoder, which have shortcomings in training efficiency and expression ability. Recently, researchers began to explore the application of transformer in image captioning. \cite{DBLP:conf/nips/HerdadeKBS19} proposed the Object Relation Transformer to introduce the region spatial information. \cite{DBLP:conf/cvpr/PanYLM20} proposed the X-Linear attention block to capture the $2^{nd}$ order interactions between the single- or multi-modal, and integrated it into the Transformer encoder and decoder. \cite{DBLP:conf/cvpr/CorniaSBC20} designed a mesh-like connectivity in decoder to exploit both low-level and high-level features from the encoder. \cite{DBLP:conf/aaai/LuoJSCWHLJ21} proposed a Dual-Level Collaborative Transformer (DLCT) to process both grid- and region-level features for realizing the complementary advantages.

Despite the outstanding performance of region-level features extracted by Faster R-CNN, the lack of fine-grained information of region-level and the time cost of Faster R-CNN pre-training are unavoidable problems. Furthermore, extracting region-level features is time-consuming, so most models directly trained and evaluated on cached features instead of image, which makes it difficult to train image captioning model end-to-end from image to descriptions.

\begin{figure*}[htb]
  \centering
  \includegraphics[width=15cm]{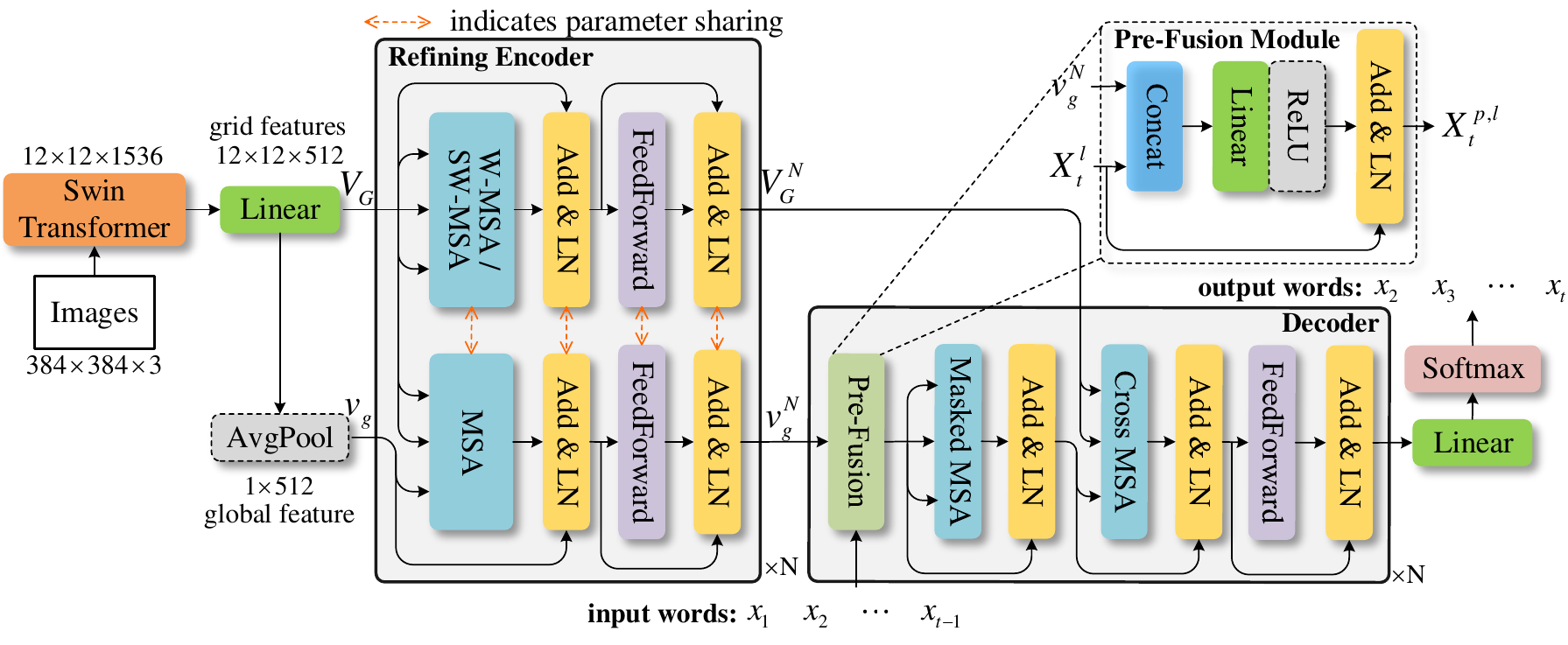}
  \caption{Overview of our proposed PureT model. We first extract image grid features $V_G$ using SwinTransformer. $v_g$ is calculated as the average pooling of $V_G$. Then $V_G$ and $v_g$ are refined into $V_G^N$ and $v_g^N$ through the Refining Encoder composed of N blocks stacks and are fed into the Decoder to generate description word by word.}\medskip
  \label{fig:overall}
\end{figure*}

\begin{figure}[htb]
  \centering
  \includegraphics[width=8.5cm]{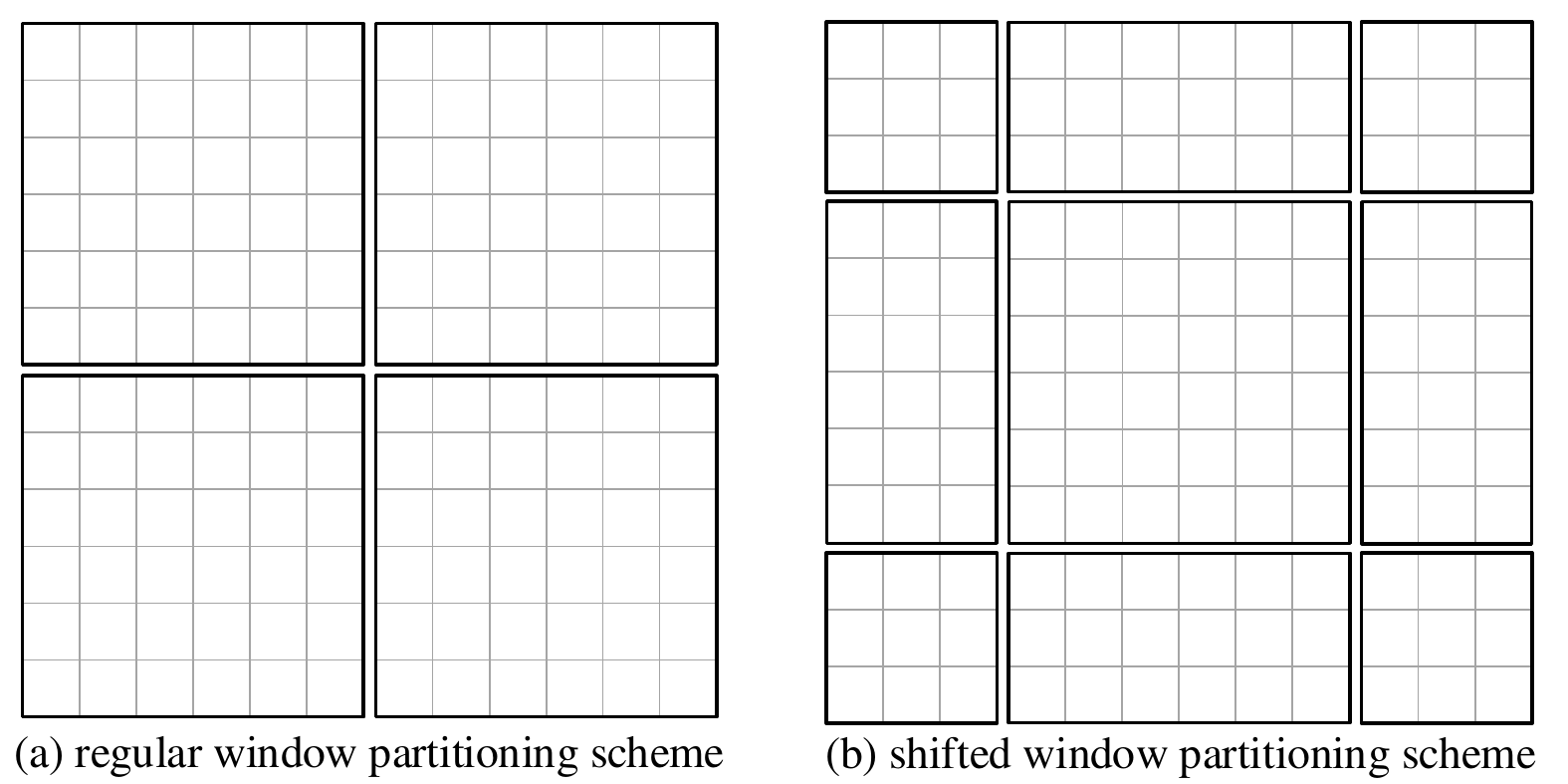}
  \caption{Illustration of regular window partitioning scheme and shifted window partitioning scheme adopted in refining encoder. The size of input feature map is $H\times W = 12\times 12$.}\medskip
  \label{fig:sw-msa}
\end{figure}

\section{Model}
The overall architecture of our PureT model is shown in Figure~\ref{fig:overall}. We adopt the widely used encoder-decoder framework, in which the encoder consists of a backbone of SwinTransformer and stacks of N refining encoder blocks and the decoder consists of stacks of N decoder blocks. The encoder is in charge of extracting grid features from the input image and refining them by capturing the intra-relationship between them. The decoder uses the refined image grid features to generate the captions word by word by capturing the inter-relationship between word and image grid features. 

\subsection{Attention Mechanism}
The attention mechanism can be abstractly summarized as follows:
\begin{equation}
\operatorname{Attention}(q, k, v) = f_{sim}(q, k)v
\end{equation}
where $f_{sim}(\cdot)$ is a function used to compute the similarity scores between some queries ($q$) and keys ($k$). The output of attention mechanism is the weighted sum on values ($v$) based on similarity scores.

In our model, Multi-head Self Attention (MSA) \cite{DBLP:conf/nips/VaswaniSPUJGKP17} and its variants Window MSA / Shifted Window MSA (W-MSA / SW-MSA) modules proposed by SwinTransformer \cite{liu2021Swin} are used, where MSA is adopted in the decoder to model the intra-relationship of word sequence and the inter-relationship between word and grid features, and W-MSA / SW-MSA are adopted in the encoder to model intra-relationship of image grid features. The above three attention modules use $\operatorname{Softmax}(\cdot)$ as the similarity scoring function, which can be formulated as follows:
\begin{equation}
\operatorname{Attention}(q, k, v) = \operatorname{Softmax}\left(\frac{qk^\mathrm{T}}{\sqrt{d_k}}\right)v
\end{equation}
where $d_k$ is the dimension of $k$. 
\begin{equation}
\begin{split}
  & \operatorname{MSA}(Q, K, V) = \operatorname{Concat}(head_1, \ldots, head_h) \\
  & head_i = \operatorname{Attention}(Q^i, K^i, V^i), i = 1,2,\ldots,h
\end{split}
\end{equation}
where $h$ is the number of heads. 
$Q^i, K^i$ and $V^i$ are the $i$-th slice of $Q, K$ and $V$ respectively, which can be formulated as follows:
\begin{align}
\bigstar = \operatorname{Concat}(\bigstar^1, \ldots, \bigstar^i, \ldots, \bigstar^h)
\end{align}
where $\bigstar \in \mathbb{R}^{L_\bigstar \times D_\bigstar}$ and $\bigstar^i \in \mathbb{R}^{L_\bigstar \times \frac{D_\bigstar}{h}}$ ($\bigstar$ refers to $Q, K$ and $V$), $L_\bigstar$ and $D_\bigstar$ are the length and dimension.

In the $i$-th head of MSA, each token of the query $Q^i$ calculates its similarity with all tokens of the key $K^i$, and performs the weighted sum on all tokens of the value $V^i$ to obtain the corresponding output. Therefore, MSA can be regarded as a global attention mechanism.

\subsubsection{W-MSA and SW-MSA} 
Aiming at the quadratic complexity caused by the global computation of MSA, SwinTransformer proposed W-MSA and SW-MSA to compute self-attention within local windows \cite{liu2021Swin}. In this paper, both W-MSA and SW-MSA are used in the encoder, in which inputs of $Q, K$ and $V$ are all from image grid features, therefore they have the same length $L = H\times W$ and dimension $D$. 

Compared with MSA, W-MSA and SW-MSA first partition the inputs of $Q, K$ and $V$ into several windows, and then apply MSA separately in each window. Figure~\ref{fig:sw-msa} illustrate the regular window partitioning scheme and shifted window partitioning scheme of W-MSA and SW-MSA respectively. Adding SW-MSA after W-MSA aims to solve the lack of connections across windows of W-MSA module to further improve the modeling ability. W-MSA and SW-MSA can be formulated as follows:
\begin{align}
    &\operatorname{(S)W-MSA}(Q, K, V) = \operatorname{Merge}(window_1, \ldots, window_w) \notag\\
    &window_i = \operatorname{MSA}(Q_W^{i}, K_W^{i}, V_W^{i}), i=1, 2, \ldots, w 
\end{align}
where $w$ is the number of windows and $\operatorname{Merge}(\cdot)$ is the reverse operation of regular/shifted window partitioning scheme. $Q_W^i, K_W^i$ and $V_W^i$ are the $i$-th window of $Q, K$ and $V$ respectively, which can be formulated as follows:
\begin{align}
\bigstar = &\operatorname{Merge}\left(\bigstar_W^1, \ldots, \bigstar_W^i, \ldots, \bigstar_W^w\right)
\end{align}
where $\bigstar \in \mathbb{R}^{L \times D}$ and $\bigstar_W^i \in \mathbb{R}^{\frac{L}{w} \times D}$ ($\bigstar$ refers to $Q, K$ and $V$).

\subsection{Encoder} 
Different from most existing models, we first employ SwinTransformer \cite{liu2021Swin} instead of pre-trained CNN or Faster R-CNN as the backbone encoder to extract a set of grid features $V_G = \{v_1, v_2, \ldots, v_m\}$ from given input images as the initial visual features, where $v_i \in \mathbb{R}^D$, $D$ is the embedding dimension of each grid feature, and $m$ is the number of grid features ($m\operatorname{=12\times 12}$ in this paper). 

After grid features $V_G$ are extracted, we refer to the standard transformer encoder \cite{DBLP:conf/nips/VaswaniSPUJGKP17} to construct a refining encoder to enhance the grid features by capturing the intra-relationship between them. Furthermore, inspired by \cite{DBLP:conf/aaai/JiLSCLW0J21}, we calculate the mean pooling of grid features $v_g = \frac{1}{m}\sum_{i=1}^mv_i$ as the initial global feature and introduce it into W-MSA and SW-MSA. Specifically, when applying MSA in each window, the global feature is added into the keys $k$ and values $v$ as an extra token. Meanwhile, we also refine the global feature by using it as an extra query $q$ token and applying MSA on all grid features.

As shown in Figure~\ref{fig:overall}, the refining encoder is composed of $N$ blocks stacked in sequence ($N=3$ in this paper), and each block consists of a W-MSA or SW-MSA module with feedforward layer, in which W-MSA and SW-MSA are used alternately. The $l$-th block can be formulated as follows:
\begin{align}
\hat{V}_G^l = &\operatorname{LayerNorm}\left(V_G^{l-1} + \left.\operatorname{(S)W-MSA}\left(W_Q^lV_G^{l-1}, \right.\right.\right. \notag\\
    &\left.\left.W_K^l\left[V_G^{l-1}; v_g^{l-1}\right]_s, W_V^l\left[V_G^{l-1}; v_g^{l-1}\right]_s\right)\right) \\
\hat{v}_g^l = &\operatorname{LayerNorm}(v_g^{l-1} + \operatorname{MSA}\left(W_Q^lv_g^{l-1}, \notag \right.\\
    &\left.\left.W_K^l[V_G^{l-1}; v_g^{l-1}]_s, W_V^l[V_G^{l-1}; v_g^{l-1}]_s\right)\right) \\
V_G^l = &\operatorname{LayerNorm}\left(\hat{V}_G^l + \operatorname{FeedForward}(\hat{V}_G^l)\right)\\
v_g^l = &\operatorname{LayerNorm}\left(\hat{v}_g^l + \operatorname{FeedForward}(\hat{v}_g^l)\right)
\end{align}
where $V_G^{l-1}$ and $v_g^{l-1}$ denote the output grid features and global feature of block $l-1$ respectively, and which are used as the input of block $l$, in which $V_G^0 = V_G$ and $v_g^0 = v_g$, $W_Q^l,Q_K^l,W_V^l\in \mathbb{R}^{D\times D}$ are learnt parameter matrices; $[V_G^{l-1} ;v_g^{l-1}]_s\in \mathbb{R}^{(k+1)\times D}$ denotes the stack operation of grid features and global feature and $\operatorname{FeedForward}(\cdot)$ consists of two linear layer with $\operatorname{ReLU}$ activation function in between, as formulated below:
\begin{align}
\operatorname{FeedForward}\left(x\right) = W_2\operatorname{ReLU}\left(W_1x\right)
\end{align}
where $W_1\in\mathbb{R}^{(4D)\times D}$ and $W_2\in\mathbb{R}^{D\times(4D)}$ are the learnt parameter matrices of two linear layers respectively. Note that the parameter of refining process for grid features and global feature are reused. The output refined grid features $V_G^N$ and refined global feature $v_g^N$ of block $N$ are fed into the decoder as the input of visual content.

\subsection{Decoder}
The decoder aims to generate the output caption word by word conditioned on the refined global and grid features from the encoder. The interaction between multi-modal occurs in this part. As shown in Figure~\ref{fig:overall}, the decoder is composed of $N$ blocks stacked in sequence ($N=3$ in this paper), where each block can be divided into three modules: 1) Pre-Fusion Module, which contains the pre-fusion process between previously generated words and refined global feature, which can be regarded as the first inter-modal interaction between natural language and visual content; 2) Language Masked MSA Module, which can be regarded as the intra-modal interaction within the generated words; 3) Cross MSA Module, which contains a MSA module with a FeedForward layer, which can be regarded as the second inter-modal interaction between visual content and natural language; 4) Word Generation Module, which contains a linear layer with softmax function.

\subsubsection{Pre-Fusion Module}
Most recent Transformer-based models only use image region or grid features without global feature, where the interaction between multi-modal features only occurs in cross attention between generated word and visual features before generating the next word. The lack of interaction of global contextual information limits reasoning capability to a certain extent. Therefore, we construct a pre-fusion module to fuse the refined global feature $v_g^N$ into the input of each block of decoder, which can be regarded as the first time multi-modal interaction to capture global visual context information and can be formulated as follows:
\begin{align}
X_{1:t-1}^{p, l} = \operatorname{Layer}&\operatorname{Norm}\left(X_{1:t-1}^{l-1} + \right.\notag\\
&\left.\operatorname{ReLU}\left(W_f\left[X_{1:t-1}^{l-1}; v_g^N\right]\right) \right)
\end{align}
where $X_{1:t-1}^{l-1}\in \mathbb{R}^{(t-1)\times D}$ denotes the output of block $l-1$ and is used as the input of block $l$ at $t$-th timestep , $\left[X_{1:t-1}^{l-1}; v_g\right]\in \mathbb{R}^{(t-1)\times 2D}$ indicates concatenation and $W_f\in \mathbb{R}^{D\times 2D}$ is learnt parameters of a linear layer; the output $X_{1:t-1}^{p,l}\in \mathbb{R}^{(t-1)\times D}$ is fed into the Language Masked MSA Module. Note that the initial input at the first block comes from the previously generated words:
\begin{align}
X_{1:t-1}^0=W_ex_{1:t-1}
\end{align}
where $x_{1:t-1}$ are one-hot encodings of the generated words before $t$-th timestep, and $W_e\in\mathbb{R}^{D\times|\Sigma|}$ is the word embedding matrix of the vocablulary $\Sigma$.

\subsubsection{Language Masked MSA Module}
The module aims to model the intra-modal relationship (words-to-words) within $X_{1:t-1}^{p,l}$, which can be formulated as follows:
\begin{align}
\tilde{X}_{t-1}^l = &\operatorname{LayerNorm}\left( X_{t-1}^{p,l} + \operatorname{MSA}\left(W_Q^{m,l}X_{t-1}^{p,l}, \right.\right.\notag\\
&\left.\left. W_K^{m,l}X_{1:t-1}^{p,l}, W_V^{m,l}X_{1:t-1}^{p,l} \right) \right)
\end{align}
where $W_Q^{m,l}, W_K^{m,l}, W_V^{m,l} \in \mathbb{R}^{D\times D}$ are learnt parameters, and $X_{t-1}^{p,l}$ indicates the corresponding embedding vector of the generated word at $(t-1)$-th timestep, which means that each word is only allowed to calculate attention map at its earlier generated words.

\begin{table*}
\begin{center}
\footnotesize
\begin{tabular}{lcccccccccccc}
\hline
\multirow{2}{*}{Models} & \multicolumn{6}{c}{Single Model} & \multicolumn{6}{c}{Ensemble Model} \\
\cmidrule(r){2-7}  \cmidrule(r){8-13}
                       & \multicolumn{1}{c}{B-1} & \multicolumn{1}{c}{B-4} & \multicolumn{1}{c}{M} & \multicolumn{1}{c}{R} & \multicolumn{1}{c}{C} & \multicolumn{1}{c}{S} & \multicolumn{1}{c}{B-1} & \multicolumn{1}{c}{B-4} & \multicolumn{1}{c}{M} & \multicolumn{1}{c}{R} & \multicolumn{1}{c}{C} & \multicolumn{1}{c}{S} \\
\hline
\multicolumn{1}{c}{} & \multicolumn{12}{c}{CNN-LSTM based models}\\
\hline
SCST  & - & 34.2 & 26.7 & 55.7 & 114.0 & - &  - & 35.4 & 27.1 & 56.6 & 117.5 & -\\
RFNet  & 79.1 & 36.5 & 27.7 & 57.3 & 121.9 & 21.2 & 80.4 & 37.9 & 28.3 & 58.3 & 125.7 & 21.7 \\
Up-Down  & 79.8 & 36.3 & 27.7 & 56.9 & 120.1 & 21.4 & - & - & - & - & - & -\\
GCN-LSTM & 80.5 &  38.2 & 28.5 & 58.3 & 127.6 & 22.0 & 80.9 & 38.3 & 28.6 & 58.5 & 128.7 & 22.1\\
AoANet  & 80.2 & 38.9 & 29.2 & 58.8 & 129.8 & 22.4 & 81.6 & 40.2 & 29.3 & 59.4 & 132.0 & 22.8\\
X-LAN & 80.8 &  39.5 & 29.5 & 59.2 & 132.0 & 23.4 & 81.6 & 40.3 & 29.8 & 59.6 & 133.7 & 23.6 \\
\hline
\multicolumn{1}{c}{} & \multicolumn{12}{c}{CNN-Transformer based models}\\
\hline
ORT & 80.5 &38.6 & 28.7 & 58.4 & 128.3 & 22.6 & - & - & - & - & - & -\\
X-Transformer & 80.9 &  39.7 & 29.5 & 59.1 & 132.8 & 23.4 & 81.7 & 40.7 & 29.9 & 59.7 & 135.3 & 23.8\\
$\mathcal{M}^2$ Transformer  & 80.8 & 39.1 & 29.2 & 58.6 & 131.2 & 22.6 & 82.0 &  40.5 & 29.7 & 59.5 & 134.5 & 23.5 \\
RSTNet & 81.8 & 40.1 & 29.8 & 59.5 & 135.6 & 23.3 & - & - & - & - & - & - \\
GET & 81.5 & 39.5 & 29.3 & 58.9 & 131.6 & 22.8 & 82.1 & 40.6 & 29.8 & 59.6 & 135.1 & 23.8 \\
DLCT & 81.4 & 39.8 & 29.5 & 59.1 & 133.8 & 23.0 & 82.2 & 40.8 & 29.9 & 59.8 & 137.5 & 23.3 \\
\hline
PureT & \textbf{82.1} & \textbf{40.9} & \textbf{30.2} & \textbf{60.1} & \textbf{138.2} & \textbf{24.2} & \textbf{83.4} & \textbf{42.1} & \textbf{30.4} & \textbf{60.8} & \textbf{141.0} & \textbf{24.3} \\
\hline
\end{tabular}
\caption{Offline evaluation results of our proposed model and other existing state-of-the-art models on MSCOCO ``Karpathy" test split, where B-$N$, M, R, C and S denote BLEU-$N$, METEOR, ROUGE-L, CIDEr and SPICE respectively.}
\label{table:offline}
\end{center}
\end{table*}

\subsubsection{Cross MSA Module}
The module aims to model the inter-modal relationship (words-to-vision) between $\tilde{X}_{1:t-1}^l$ and $V_G^N$, which can be regarded as the second time multi-modal interaction to capture local visual context information and can be formulated as follows:
\begin{align}
\hat{X}_{t-1}^l = &\operatorname{LayerNorm}\left( \tilde{X}_{t-1}^l + \operatorname{MSA}\left( W_Q^{c,l}\tilde{X}_{t-1}^l, \right.\right.\notag\\
&\left.\left.W_K^{c,l}V_G^N, W_V^{c,l}V_G^N\right) \right) \\
X_{t-1}^l = &\operatorname{LayerNorm}(\hat{X}_{t-1}^l + \operatorname{FeedForward}(\hat{X}_{t-1}^l) )
\end{align}
where $W_Q^{c,l}, W_K^{c,l}, W_V^{c,l} \in \mathbb{R}^{D\times D}$ are learnt parameters, $\tilde{X}_{t-1}^l$ from the Language Masked MSA Module is fed into MSA as query, and refined grid features $V_G^N$ from the last block of encoder are fed into MSA as keys and values.

\subsubsection{Word Generation Module}
Given the output $X_{1:t-1}^N$ of the last decoder block, the conditional distribution over the vocablary $\Sigma$ is given by:
\begin{align}
p(x_t|x_{1:t-1}) = \operatorname{Softmax}(W_xX_{t-1}^N)
\end{align}
where $W_x\in\mathbb{R}^{|\Sigma|\times D}$ is learnt parameters. 

\subsection{Objective Functions}
We first optimize our model by applying cross entropy (XE) loss as the objective function:
\begin{equation}
L_{XE}(\theta)=-\sum_{t=1}^T\log(p_\theta(y_t^*|y_{1:t-1}^*))
\end{equation}
where $y_{1:T}^*$ is the target ground truth sequence, and $\theta$ denotes the parameters of our model. Then, we adopt self-critical sequence training (SCST) strategy \cite{DBLP:conf/cvpr/RennieMMRG17} to optimize CIDEr \cite{DBLP:conf/cvpr/VedantamZP15} metrics:
\begin{equation}
L_R(\theta)=-\operatorname{\textbf{E}}_{y_{1:T}\sim p_\theta}[r(y_{1:T})]
\end{equation}
where $r(\cdot)$ is the score of CIDEr. The gradient of $L_R$ can be approximated as follows:
\begin{equation}
\nabla_\theta L_R(\theta)\approx -\left(r(y_{1:T}^s)-r(\hat{y}_{1:T})\right)\nabla_\theta \log p_\theta(y_{1:T}^s)
\end{equation}
where $y_{1:T}^s$ is a sampled caption and $r(\hat{y}_{1:T}^s)$ defines the greedily decoded score obtained from the current model.

\begin{table*}
\begin{center}
\footnotesize
\begin{tabular}{lcccccccccccccc}
\hline
\multirow{2}{*}{Models} & \multicolumn{2}{c}{BLEU-1} & \multicolumn{2}{c}{BLEU-2} & \multicolumn{2}{c}{BLEU-3} & \multicolumn{2}{c}{BLEU-4} & \multicolumn{2}{c}{METEOR} & \multicolumn{2}{c}{ROUGE-L} & \multicolumn{2}{c}{CIDEr} \\ 
\cmidrule(r){2-3}  \cmidrule(r){4-5} \cmidrule(r){6-7} \cmidrule(r){8-9} \cmidrule(r){10-11} \cmidrule(r){12-13} \cmidrule(r){14-15}
                       & \multicolumn{1}{c}{c5} & \multicolumn{1}{c}{c40} & \multicolumn{1}{c}{c5} & \multicolumn{1}{c}{c40} & \multicolumn{1}{c}{c5} & \multicolumn{1}{c}{c40} & \multicolumn{1}{c}{c5} & \multicolumn{1}{c}{c40} & \multicolumn{1}{c}{c5} & \multicolumn{1}{c}{c40} & \multicolumn{1}{c}{c5} & \multicolumn{1}{c}{c40} & \multicolumn{1}{c}{c5} & \multicolumn{1}{c}{c40} \\ 
\hline
SCST & 78.1 & 93.7 & 61.9 & 86.0 & 47.0 & 75.9 & 35.2 & 64.5 & 27.0 & 35.5 & 56.3 & 70.7 & 114.7 & 116.7 \\ 
GCN-LSTM & 80.8 & 95.2 & 65.5 & 89.3 & 50.8 & 80.3 & 38.7 & 69.7 & 28.5 & 37.6 & 58.5 & 73.4 & 125.3 & 126.5 \\
Up-Down & 80.2 & 95.2 & 64.1 & 88.8 & 49.1 & 79.4 & 36.9 & 68.5 & 27.6 & 36.7 & 57.1 & 72.4 & 117.9 & 120.5 \\
SGAE  & 81.0 & 95.3 & 65.6 & 89.5 & 50.7 & 80.4 & 38.5 & 69.7 & 28.2 & 37.2 & 58.6 & 73.6 & 123.8 & 126.5 \\
AoANet & 81.0 & 95.0 & 65.8 & 89.6 & 51.4 & 81.3 & 39.4 & 71.2 & 29.1 & 38.5 & 58.9 & 74.5 & 126.9 & 129.6\\
X-Transformer  & 81.9 & 95.7 & 66.9 & 90.5 & 52.4 & 82.5 & 40.3 & 72.4 & 29.6 & 39.2 & 59.5 & 75.0 & 131.1 & 133.5 \\
$\mathcal{M}^2$ Transformer & 81.6 & 96.0 & 66.4 & 90.8 & 51.8 & 82.7 & 39.7 & 72.8 & 29.4 & 39.0 & 59.2 & 74.8 & 129.3 & 132.1 \\
RSTNet & 82.1 & 96.4 & 67.0 & 91.3 & 52.2 & 83.0 & 40.0 & 73.1 & 29.6 & 39.1 & 59.5 & 74.6 & 131.9 & 134.0 \\
GET & 81.6 & 96.1 & 66.5 & 90.9 & 51.9 & 82.8 & 39.7 & 72.9 & 29.4 & 38.8 & 59.1 & 74.4 & 130.3 & 132.5 \\
DLCT & 82.4 & \textbf{96.6} & 67.4 & 91.7 & 52.8 & 83.8 & 40.6 & 74.0 & 29.8 & 39.6 & 59.8 & 75.3 & 133.3 & 135.4 \\
\hline
PureT & \textbf{82.8} & 96.5 & \textbf{68.1} & \textbf{91.8} & \textbf{53.6} & \textbf{83.9} & \textbf{41.4} & \textbf{74.1} & \textbf{30.1} & \textbf{39.9} & \textbf{60.4} & \textbf{75.9} & \textbf{136.0} & \textbf{138.3} \\
\hline
\end{tabular}
\caption{Online evaluation results of our proposed model and other existing state-of-the-art models on MSCOCO.}
\label{table:online}
\end{center}
\end{table*}

\section{Experiments}

\subsection{Dataset and Evaluation Metrics}
We conduct experiments on the MSCOCO 2014 dataset \cite{DBLP:conf/eccv/LinMBHPRDZ14}, which contains 123287 images (82783 for training and 40504 for validation), and each is annotated with 5 reference captions. In this paper, we follow the ``Karpathy'' split \cite{DBLP:journals/pami/KarpathyF17} to redivide the MSCOCO, where 113287 images for training, 5000 images for validation and 5000 images for offline evaluation. Besides, MSCOCO also provides 40775 images for online testing. For the training process, we convert all training captions to lower case and drop the words occur less than 6 times, collect the rest 9487 words as our vocabulary $\Sigma$.

For fair evaluation, we adopt five widely used metrics to evaluate the quality of generated captions, including BLEU \cite{DBLP:conf/acl/PapineniRWZ02}, METEOR \cite{DBLP:conf/wmt/LavieA07}, ROUGE-L \cite{lin-2004-rouge}, CIDEr \cite{DBLP:conf/cvpr/VedantamZP15}, and SPICE \cite{DBLP:conf/eccv/AndersonFJG16}. 

\subsection{Experimental Settings}
We set the model embedding size $D$ to 512, the number of transformer heads to 8, the number of blocks $N$ for both refining encoder and decoder to 3. For the training process, we first train our model under XE loss $L_{XE}$ for 20 epochs, and set the batch size to 10 and warmup steps to 10,000; then we train our model under $L_R$ for another 30 epochs with fixed learning rate of $5\times 10^{-6}$. We adopt Adam \cite{DBLP:journals/corr/KingmaB14} optimizer in both above stages and the beam size is set to 5 in validation and evaluation process.

\subsection{Comparisons with State-of-The-Art Models}
\subsubsection{Offline Evaluation}
Table~\ref{table:offline} reports the performances of some existing state-of-the-art models and our proposed model on MSCOCO offline test split. The compared models include: 
SCST \cite{DBLP:conf/cvpr/RennieMMRG17}, 
RFNet \cite{DBLP:conf/eccv/JiangMJLZ18}, 
Up-Down \cite{DBLP:conf/cvpr/00010BT0GZ18}, 
GCN-LSTM \cite{DBLP:conf/eccv/YaoPLM18}, 
AoANet \cite{huang2019attention} and X-LAN \cite{DBLP:conf/cvpr/PanYLM20}; 
ORT \cite{DBLP:conf/nips/HerdadeKBS19}, 
X-Transformer \cite{DBLP:conf/cvpr/PanYLM20}, 
$\mathcal{M}^2$ Transformer \cite{DBLP:conf/cvpr/CorniaSBC20}, 
RSTNet\cite{DBLP:conf/cvpr/ZhangSLJZWHJ21},
GET \cite{DBLP:conf/aaai/JiLSCLW0J21} and DLCT \cite{DBLP:conf/aaai/LuoJSCWHLJ21}. We divide these models into CNN-LSTM based models and CNN-Transformer based models according to the difference mothods adopted in decoder.

For fair comparisons, we report the results of a single model and ensemble of 4 models after SCST training. As shown in Table~\ref{table:offline}, both our single model and ensemble of 4 models achieve best performances in all metrics. In the case of single model, the CIDEr score of our model reaches 138.2\%, which achieves advancements of 2.6\% and 4.4\% to the strong competitors RSTNet and DLCT. Meanwhile, our model achieves improvements of over 0.6\% to RSTNet, and improvements of over 1.0\% to DLCT in terms of metrics BLEU-4, ROUGE-L and SPICE. In the case of ensemble model, our model also achieves the best performance, and advances all other models by more than 1.0\% over all metrics except METEOR. In particular, the CIDEr score of our ensemble model reaches 141.0\%, which achieves advancements of 3.5\% and 5.9\% to DLCT and GET.

In general, the significant improvements of all metrics (especially CIDEr) demonstrate the advantage of our proposed model. In addition, compared to models that use region-level features or both region and grid-level features, our model has a relatively more balanced computational cost due to it avoids the prediction of object regions coordinates. And our model can be trained end-to-end, which allows us to explore it in more actual scenes.

\subsubsection{Online Evaluation}
As shown in Table~\ref{table:online}, we also report the performance with 5 reference captions (c5) and 40 reference captions (c40) of our model on the MSCOCO official online test server. Compared to the other state-of-the-arts, our model achieves the best scores in all metrics except a slightly lower 0.1\% in BLEU-1 (c40) than DLCT. Notably, the scores of CIDEr (c5) and CIDEr (c40) of our model reach 136.0\% and 138.3\%, which achieve advancements of 2.7\% and 2.9\% with respect to the best performer DLCT.

\begin{figure}[htb]
  \centering
  \includegraphics[width=8.5cm]{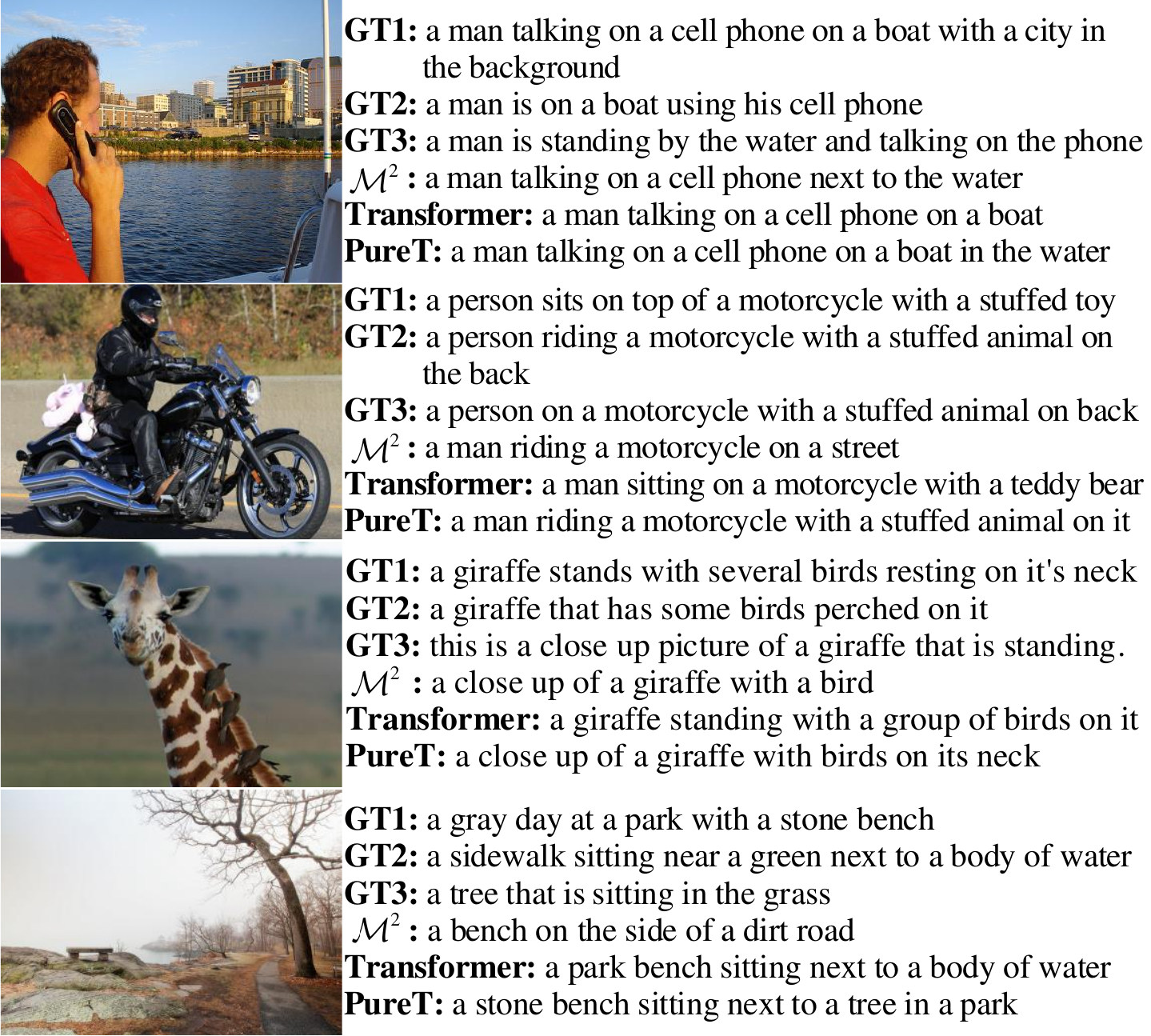}
  \caption{Examples of captions generated by standard Transformer, $\mathcal{M}^2$ Transformer and our PureT with ground-truths.}\medskip
  \label{fig:cap_example}
\end{figure}

\begin{figure*}[htb]
  \centering
  \includegraphics[width=17cm]{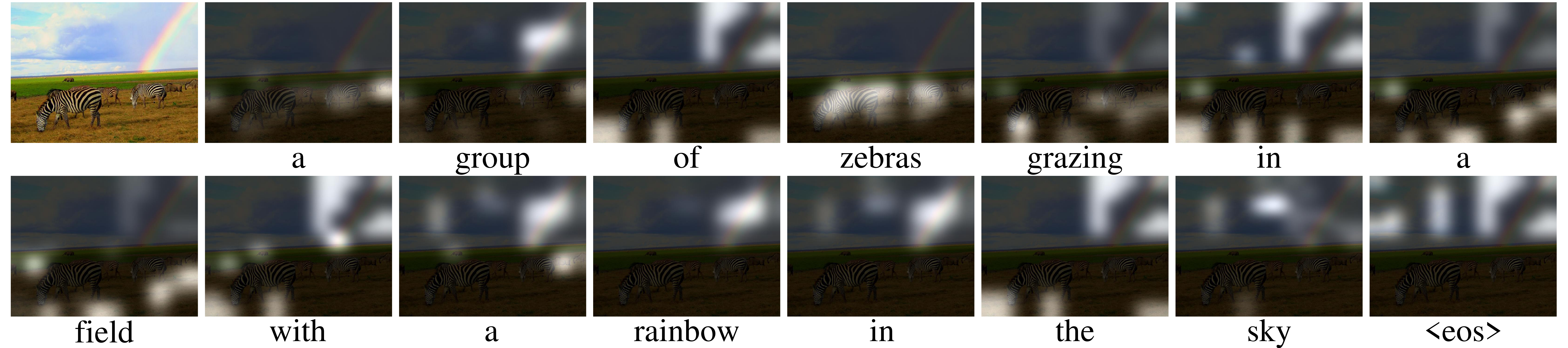}
  \caption{Visualization of attention heatmap on image along caption generation process. For each generated word, we show the image with different brigtness to represent the difference of attention weights.}\medskip
  \label{fig:att_vis}
\end{figure*}

\subsection{Ablation Study}
We conduct several ablation studies to quantify the influences of different modules in our model.

\subsubsection{Influence of W-MSA and SW-MSA} 
For quantifying the influence of W-MSA and SW-MSA in our Refining Encoder, we ablate our model with different configurations of window size $ws$ and shift size $ss$ as shown in Table~\ref{table:ablation_1}. The number of refining encoder and decoder blocks is set to 3. Note that the input $V_G\in \mathbb{R}^{m\times D}$ of Refining Encoder has a size of $m\operatorname{=12\times 12}$ in this paper. The W-MSA and SW-MSA degenerate into MSA when $ws\operatorname{=12}$ and SW-MSA into W-MSA when $ss\operatorname{=0}$. It can be seen that the model with only MSA ($ws\operatorname{=12},ss\operatorname{=0}$) performs better than the model with only W-MSA ($ws\operatorname{=6},ss\operatorname{=0}$) because W-MSA lacks connections across windows. However, the model combining W-MSA and SW-MSA ($ws\operatorname{=6},ss\operatorname{=3}$) can improve the performance of both models above in all metrics.

\begin{table}
\begin{center}
\footnotesize
\begin{tabular}{rrcccccc}
\hline
$ws$ & $ss$ & B-1 & B-4 & M & R & C & S \\
\hline
12  & 0 & 82.0 & 40.3 & 29.9 & 59.9 & 137.5 & 23.8  \\
6   & 0 & 81.8 & 40.1 & 29.9 & 59.7 & 136.8 & 23.8  \\
6   & 3 & \textbf{82.1} & \textbf{40.9} & \textbf{30.2} & \textbf{60.1} & \textbf{138.2} & \textbf{24.2}  \\
\hline
\end{tabular}
\caption{Performance comparison of different configuration of window size $ws$ and shift size $ss$.}
\label{table:ablation_1}
\end{center}
\end{table}

\subsubsection{Influence of Pre-Fusion module}
\begin{table}
\begin{center}
\footnotesize
\begin{tabular}{lcccccc}
\hline
Models  & B-1 & B-4 & M & R & C & S \\
\hline
Transformer & 81.6 & 39.8 & 29.9 & 59.6 & 136.4 & 23.8 \\
\makecell[l]{Transformer \\\quad\quad + p-f.}  &  82.0 & 40.3 & 29.9 & 59.9 & 137.5 & 23.8 \\
\hline
PureT (w/o p-f.)  & 81.8 & 40.3 & 30.0 & 59.9 & 137.9 & 24.0  \\
PureT & \textbf{82.1} & \textbf{40.9} & \textbf{30.2} & \textbf{60.1} & \textbf{138.2} & \textbf{24.2} \\
\hline
\end{tabular}
\caption{Performance comparison with / without Pre-Fusion for standard Transformer and our proposed PureT.}
\label{table:ablation_2}
\end{center}
\end{table}
To demonstrate the effectiveness of the Pre-Fusion module in our Decoder, we remove the Pre-Fusion module from our PureT model and compare it with the full model as shown in rows 4 and 5 of Table~\ref{table:ablation_2}. It can be seen that the Pre-Fusion module improves the performance in all metrics. Furthermore, we construct the standard Transformer (3 blocks of encoder/decoder) as the baseline model, which reaches an excellent performance as shown in row 1 in Table~\ref{table:ablation_2}. Then we extend the baseline model by adding the Pre-Fusion module (equivalent to the model in row 1 of Table~\ref{table:ablation_1}), which also has a better performance in all metrics.

\subsubsection{Influence of the number of stacked blocks}
We also conduct several experiments to evaluate the influence of the number of the Refining Encoder and Decoder blocks. As shown in Table~\ref{table:ablation_3}, models with more than 2 blocks have a significant improvement (more than 2.0\%) in CIDEr score compare to the model with 1 block. Note that the model with 4 blocks has a significant advantage in BLEU scores to other models, but considering the increase of model parameters and the sufficiently excellent performance of the model with 3 blocks, we set the number of blocks $N$ to 3 as the final configuration. Remarkably, the model with only 1 block also has a better performance in comparison to earlier state-of-the-art works (e.g. RSTNet, GET and DLCT) in Table~\ref{table:offline}, which further indicates the effectiveness of our model.

\begin{table}
\begin{center}
\footnotesize
\begin{tabular}{ccccccc}
\hline
Layer & B-1 & B-4 & M & R & C & S \\
\hline
1 & 81.8 & 40.2 & 29.7 & 59.5 & 135.8 & 23.5  \\
2 & 81.8 & 40.5 & 30.0 & 59.9 & \textbf{138.2} & 23.9  \\
3 & 82.1 & 40.9 & \textbf{30.2} & \textbf{60.1} & \textbf{138.2} & \textbf{24.2}  \\
4 & \textbf{82.7} & \textbf{41.1} & 30.0 & \textbf{60.1} & \textbf{138.2} & 24.0  \\
\hline
\end{tabular}
\caption{Performance comparison of different number of Refining Encoder and Decoder blocks.}
\label{table:ablation_3}
\end{center}
\end{table}

\subsubsection{Influence of different backbone}

\begin{table*}
\setcounter{table}{5}
\begin{center}
\footnotesize
\begin{tabular}{lllclcccccccc}
\hline
Baseline Models & Backbone & Feat. Type & Feat. Size & N & B-1 & B-2 & B-3 & B-4 & M & R & C & S \\
\hline
\multirow{3}{*}{$\mathcal{M}^2$ Transformer} & ResNet-101 & Region & (10$-$100) & 3$^\dagger$ & 80.8 & - & - & 39.1 & 29.2 & 58.6 & 131.2 & 22.6 \\
 & ResNeXt-101 & Grid & $7\times 7$ & 3$^\ddagger$ & 80.8 & - & - & 38.9 & 29.1 & 58.5 & 131.7 & 22.6  \\
 & SwinTransformer & Grid & $12\times 12$ & 3 & 81.8 & 66.8 & \textbf{52.6} & 40.5 & 29.6 & 59.9 & 135.4 & 23.3 \\
\hline
\multirow{4}{*}{X-Transformer} & ResNet-101 & Region & (10$-$100) & 6$^\dagger$ & 80.9 & 65.8 & 51.5 & 39.7 & 29.5 & 59.1 & 132.8 & 23.4 \\
 & ResNeXt-101 & Grid & $7\times 7$ & 6$^\ddagger$ & 81.0 & - & - & 39.7 & 29.4 & 58.9 & 132.5 & 23.1  \\
 & SwinTransformer & Grid & $12\times 12$ & 6 &  81.4 & 66.3 & 52.0 & 39.9 & 29.5 & 59.5 & 133.7 & 23.4 \\
 & SwinTransformer & Grid & $12\times 12$ & 3 &  81.9 & 66.7 & 52.3 & 40.1 & 29.6 & 59.6 & 134.8 & 23.4 \\
\hline
\multirow{4}{*}{\makecell[l]{standard \\Transformer}} & ResNet-101 & Region & (10$-$100) & 3 & 80.0 & 64.9 & 50.5 & 38.7 & 29.0 & 58.6 & 130.1 & 22.9 \\
 & ResNeXt-101 & Grid & $7\times 7$ & 3$^\ddagger$ & 81.2 & - & - & 39.0 & 29.2 & 58.9 & 131.7 & 22.6 \\
 & ResNeXt-101 & Grid & $12\times 12$ & 3 & 80.8 & 65.8 & 51.4 & 39.4 & 29.4 & 59.2 & 132.8 & 23.2 \\
 & SwinTransformer & Grid & $12\times 12$ & 3 & 81.6 & 66.5 & 52.0 & 39.8 & 29.9 & 59.6 & 136.4 & 23.8 \\
\hline
\multirow{2}{*}{PureT} & ResNeXt-101 & Grid & $12\times 12$ & 3 & 80.7 & 65.9 & 51.7 & 39.9 & 29.2 & 59.1 & 131.8 & 23.0 \\
 & ViT & Grid & $12\times 12$ & 3 & 81.6 & 66.6 & 52.3 & 40.3 & 29.7 & 59.5 & 135.2 & 23.6 \\
 & SwinTransformer & Grid & $12\times 12$ & 3 & \textbf{82.1} &  \textbf{67.3} & 52.0 & \textbf{40.9} & \textbf{30.2} & \textbf{60.1} & \textbf{138.2} & \textbf{24.2} \\
\hline
\end{tabular}
\caption{Performance comparison of different configuration of backbone models. ResNet-101 and ResNeXt-101 indicate Faster R-CNN in conjunction with them respectively. Region features extracted by ResNet-101 have adaptive size of 10 to 100. Grid features extracted by ResNeXt-101 can be extracted in the size of $12\times 12$ or $7\times 7$ by average pooling as need. Grid features (SwinTransformer) are extracted in the size of $12\times 12$. $N$ denotes the number of encoder and decoder blocks, superscript $\dagger$ indicates that the results are from the respectively official paper and $\ddagger$ indicates that the results are from \cite{DBLP:conf/aaai/LuoJSCWHLJ21}, and other results come from our experiments.}
\label{table:ablation_exp}
\end{center}
\end{table*}

To quantify the influence of different features extracted by different backbone models, we adopt different image captioning models, as baseline models and ablate them with different configurations of backbone models as shown in Table~\ref{table:ablation_exp}. The baseline models include: $\mathcal{M}^2$ Transformer \cite{DBLP:conf/cvpr/CorniaSBC20}, X-Transformer \cite{DBLP:conf/cvpr/PanYLM20} and standard Transformer \cite{DBLP:conf/nips/VaswaniSPUJGKP17}. The backbone models include: Faster R-CNN \cite{DBLP:conf/nips/RenHGS15} in conjunction with ResNet-101, which is adopted in \cite{DBLP:conf/cvpr/00010BT0GZ18}; Faster R-CNN in conjunction with ResNeXt-101, which is adopted in \cite{DBLP:conf/cvpr/JiangMRLC20}; ViT \cite{DBLP:conf/iclr/DosovitskiyB0WZ21} and SwinTransformer \cite{liu2021Swin}. 

As we can see, grid features extracted by SwinTransformer can achieve significant performance improvement compared with region features extracted by ResNet-101 and grid features extracted by ResNeXt-101 and ViT.

In terms of $\mathcal{M}^2$ Transformer and X-Transformer, the backbone models of ResNet-101 and ResNeXt-101 have similar performance. The backbone model of SwinTransformer comprehensively improves scores of all metrics, which boosts the CIDEr score more than 3.7\% in $\mathcal{M}^2$ Transformer especially. Note that the backbone model with $N=3$ has a better performance than $N=6$ in X-Transformer, which indicates the superiority of SwinTransformer in image captioning and allows us to explore more tiny and efficient models and apply it in more actual scenes. In terms of standard Transformer, the backbone model of SwinTransformer reaches an excellent performance and is even better than $\mathcal{M}^2$ Transformer and X-Transformer in scores of METEOR, CIDEr and SPICE. In terms of our PureT, the backbone of SwinTransformer also achieves a better performance than ResNeXt-101.

In general, in our extensive experiments, we find that the backbone models of CNN (e.g. Faster RCNN in conjunction with ResNet-101 or ResNeXt-101) are more suitable for using LSTM or Transformer with non-standard MSA (e.g. X-Transformer) as decoder, and the backbone of SwinTransformer is more suitable for using Transformer with standard MSA (e.g. $\mathcal{M}^2$ Transformer, standard Transformer and our PureT) as decoder. Therefore, we intend to explore a lighter and simpler Transformer-based model in our future work.

\subsubsection{Influence of different Refining Encoder}
\begin{table}
\begin{center}
\footnotesize
\begin{tabular}{ccccccc}
\hline
Ref. Enc. & B-1 & B-4 & M & R & C & S \\
\hline
w/o            & 81.5 & 39.5 & 29.3 & 59.2 & 134.3 & 23.0  \\
$\mathcal{M}^2$ & 81.9 & 40.2 & 29.6 & 59.7 & 135.9 & 23.7  \\
X              & 81.7 & 40.0 & 29.7 & 59.5 & 135.5 & 23.5  \\
PureT & \textbf{82.1} & \textbf{40.9} & \textbf{30.2} & \textbf{60.1} & \textbf{138.2} & \textbf{24.2}  \\
\hline
\end{tabular}
\caption{Performance comparison of different Refining Encoder. w/o indicates deleting Refining Encoder, $\mathcal{M}^2$ and X indicate replacing Refining Encoder with encoders of $\mathcal{M}^2$ Transformer and X-Transformer respectively.}
\label{table:ablation_exp_1}
\end{center}
\end{table}
To further quantify the influence of Refining Encoder, we ablate the Refining Encoder by different configurations as shown in Table~\ref{table:ablation_exp_1}. We delete the Refining Encoder to confirm whether the Refining Encoder is a necessary module, and replace our proposed Refining Encoder with encoders of $\mathcal{M}^2$ Transformer and X-Transformer to verify the advantages of our Refining Encoder. As we can see, deleting Refining Encoder can also achieve good performance, which is better than most existing SOTAs in Table~\ref{table:offline}. But our proposed Refining Encoder or other encoders bring significant performance gain than deleting Refining Encoder, which denotes the importance of Refining Encoder. Our proposed Refining Encoder brings the maximum gain and achieves the best performance than other, which denotes that the effectiveness and advantages of our proposed Refining Encoder.

\subsection{Visualization Analysis}
Figure~\ref{fig:cap_example} proposes some example image captions generated by $\mathcal{M}^2$ Transformer (official model), standard Transformer and our PureT. Note that $\mathcal{M}^2$ Transformer adopts Faster R-CNN, standard Transformer and PureT adopt SwinTransformer as the encoder. Generally, our PureT is able to catch additional fine-grained information and generate more accurate and descriptive captions.

To qualitatively evaluate the effect of our PureT, we give the visualization of attention heatmap on the image along caption generation process in Figure~\ref{fig:att_vis}. It can be observed that our model can attend to correct areas when generating words. When generating nominal words, such as ``zebras'', ``rainbow'', ``field'' and ``sky'', the attention heatmap is correctly transformed into the body area of the corresponding objects. In addition, our model focuses on the nearby areas of zebra heads when generating ``grazing'', which correctly captures the semantic information and confirms the advantages of our model.

\section{Conclusion}
In this paper, we propose a pure Transformer-based model, which adopts SwinTransformer as the backbone encoder and can be trained end-to-end from image to descriptions easily. Furthermore, we construct a refining encoder to refine both image grid features and global feature with the mutual guidance between them, which realizes the complementary advantages between local and global attention. We also fuse the refined global feature with previously generated words in the decoder to enhance the multi-modal interaction, which further improves the modeling capability. Experimental results on MSCOCO dataset demonstrate that our proposed model achieves a new state-of-the-art performance. 

\bibliography{aaai22}

\begin{thebibliography}{36}
\providecommand{\natexlab}[1]{#1}

\bibitem[{Anderson et~al.(2016)Anderson, Fernando, Johnson, and
  Gould}]{DBLP:conf/eccv/AndersonFJG16}
Anderson, P.; Fernando, B.; Johnson, M.; and Gould, S. 2016.
\newblock {SPICE:} Semantic Propositional Image Caption Evaluation.
\newblock In \emph{Proceedings of the ECCV}, 382--398.

\bibitem[{Anderson et~al.(2018)Anderson, He, Buehler, Teney, Johnson, Gould,
  and Zhang}]{DBLP:conf/cvpr/00010BT0GZ18}
Anderson, P.; He, X.; Buehler, C.; Teney, D.; Johnson, M.; Gould, S.; and
  Zhang, L. 2018.
\newblock Bottom-Up and Top-Down Attention for Image Captioning and Visual
  Question Answering.
\newblock In \emph{Proceedings of the CVPR}, 6077--6086.

\bibitem[{Cho et~al.(2014)Cho, van Merrienboer, G{\"{u}}l{\c{c}}ehre, Bahdanau,
  Bougares, Schwenk, and Bengio}]{DBLP:conf/emnlp/ChoMGBBSB14}
Cho, K.; van Merrienboer, B.; G{\"{u}}l{\c{c}}ehre, {\c{C}}.; Bahdanau, D.;
  Bougares, F.; Schwenk, H.; and Bengio, Y. 2014.
\newblock Learning Phrase Representations using {RNN} Encoder-Decoder for
  Statistical Machine Translation.
\newblock In \emph{Proceedings of the EMNLP}, 1724--1734.

\bibitem[{Cornia et~al.(2020)Cornia, Stefanini, Baraldi, and
  Cucchiara}]{DBLP:conf/cvpr/CorniaSBC20}
Cornia, M.; Stefanini, M.; Baraldi, L.; and Cucchiara, R. 2020.
\newblock Meshed-Memory Transformer for Image Captioning.
\newblock In \emph{Proceedings of the CVPR}, 10575--10584.

\bibitem[{Dosovitskiy et~al.(2021)Dosovitskiy, Beyer, Kolesnikov, Weissenborn,
  Zhai, Unterthiner, Dehghani, Minderer, Heigold, Gelly, Uszkoreit, and
  Houlsby}]{DBLP:conf/iclr/DosovitskiyB0WZ21}
Dosovitskiy, A.; Beyer, L.; Kolesnikov, A.; Weissenborn, D.; Zhai, X.;
  Unterthiner, T.; Dehghani, M.; Minderer, M.; Heigold, G.; Gelly, S.;
  Uszkoreit, J.; and Houlsby, N. 2021.
\newblock An Image is Worth 16x16 Words: Transformers for Image Recognition at
  Scale.
\newblock In \emph{Proceedings of the ICLR}. OpenReview.net.

\bibitem[{He et~al.(2016)He, Zhang, Ren, and Sun}]{DBLP:conf/cvpr/HeZRS16}
He, K.; Zhang, X.; Ren, S.; and Sun, J. 2016.
\newblock Deep Residual Learning for Image Recognition.
\newblock In \emph{Proceedings of the CVPR}, 770--778.

\bibitem[{Herdade et~al.(2019)Herdade, Kappeler, Boakye, and
  Soares}]{DBLP:conf/nips/HerdadeKBS19}
Herdade, S.; Kappeler, A.; Boakye, K.; and Soares, J. 2019.
\newblock Image Captioning: Transforming Objects into Words.
\newblock In \emph{Proceedings of the NeurIPS}, 11135--11145.

\bibitem[{Hochreiter and Schmidhuber(1997)}]{DBLP:journals/neco/HochreiterS97}
Hochreiter, S.; and Schmidhuber, J. 1997.
\newblock Long Short-Term Memory.
\newblock \emph{Neural Computation}, 9(8): 1735--1780.

\bibitem[{Huang et~al.(2019)Huang, Wang, Chen, and Wei}]{huang2019attention}
Huang, L.; Wang, W.; Chen, J.; and Wei, X. 2019.
\newblock Attention on Attention for Image Captioning.
\newblock In \emph{Proceedings of the ICCV}, 4633--4642.

\bibitem[{Ji et~al.(2021)Ji, Luo, Sun, Chen, Luo, Wu, Gao, and
  Ji}]{DBLP:conf/aaai/JiLSCLW0J21}
Ji, J.; Luo, Y.; Sun, X.; Chen, F.; Luo, G.; Wu, Y.; Gao, Y.; and Ji, R. 2021.
\newblock Improving Image Captioning by Leveraging Intra- and Inter-layer
  Global Representation in Transformer Network.
\newblock In \emph{Proceedings of the AAAI}, 1655--1663. {AAAI} Press.

\bibitem[{Jiang et~al.(2020)Jiang, Misra, Rohrbach, Learned{-}Miller, and
  Chen}]{DBLP:conf/cvpr/JiangMRLC20}
Jiang, H.; Misra, I.; Rohrbach, M.; Learned{-}Miller, E.~G.; and Chen, X. 2020.
\newblock In Defense of Grid Features for Visual Question Answering.
\newblock In \emph{Proceedings of the CVPR}, 10264--10273. {IEEE}.

\bibitem[{Jiang et~al.(2018)Jiang, Ma, Jiang, Liu, and
  Zhang}]{DBLP:conf/eccv/JiangMJLZ18}
Jiang, W.; Ma, L.; Jiang, Y.; Liu, W.; and Zhang, T. 2018.
\newblock Recurrent Fusion Network for Image Captioning.
\newblock In \emph{Proceedings of the ECCV}, 510--526.

\bibitem[{Karpathy and Fei{-}Fei(2017)}]{DBLP:journals/pami/KarpathyF17}
Karpathy, A.; and Fei{-}Fei, L. 2017.
\newblock Deep Visual-Semantic Alignments for Generating Image Descriptions.
\newblock \emph{{IEEE} Trans. Pattern Anal. Mach. Intell.}, 39(4): 664--676.

\bibitem[{Kingma and Ba(2015)}]{DBLP:journals/corr/KingmaB14}
Kingma, D.~P.; and Ba, J. 2015.
\newblock Adam: {A} Method for Stochastic Optimization.
\newblock In \emph{Proceedings of the ICLR}.

\bibitem[{Krishna et~al.(2017)Krishna, Zhu, Groth, Johnson, Hata, Kravitz,
  Chen, Kalantidis, Li, Shamma, Bernstein, and
  Fei{-}Fei}]{DBLP:journals/ijcv/KrishnaZGJHKCKL17}
Krishna, R.; Zhu, Y.; Groth, O.; Johnson, J.; Hata, K.; Kravitz, J.; Chen, S.;
  Kalantidis, Y.; Li, L.; Shamma, D.~A.; Bernstein, M.~S.; and Fei{-}Fei, L.
  2017.
\newblock Visual Genome: Connecting Language and Vision Using Crowdsourced
  Dense Image Annotations.
\newblock \emph{IJCV}, 123(1): 32--73.

\bibitem[{Lavie and Agarwal(2007)}]{DBLP:conf/wmt/LavieA07}
Lavie, A.; and Agarwal, A. 2007.
\newblock {METEOR:} An Automatic Metric for {MT} Evaluation with High Levels of
  Correlation with Human Judgments.
\newblock In \emph{Proceedings of the ACL}, 228--231.

\bibitem[{Li et~al.(2019)Li, Zhu, Liu, and Yang}]{GuangLi2019}
Li, G.; Zhu, L.; Liu, P.; and Yang, Y. 2019.
\newblock Entangled Transformer for Image Captioning.
\newblock In \emph{Proceedings of the ICCV}, 8927--8936.

\bibitem[{Lin(2004)}]{lin-2004-rouge}
Lin, C.-Y. 2004.
\newblock {ROUGE}: A Package for Automatic Evaluation of Summaries.
\newblock In \emph{Proceedings of the ACL-04 workshop on Text Summarization
  Branches Out}.

\bibitem[{Lin et~al.(2014)Lin, Maire, Belongie, Hays, Perona, Ramanan,
  Doll{\'{a}}r, and Zitnick}]{DBLP:conf/eccv/LinMBHPRDZ14}
Lin, T.; Maire, M.; Belongie, S.~J.; Hays, J.; Perona, P.; Ramanan, D.;
  Doll{\'{a}}r, P.; and Zitnick, C.~L. 2014.
\newblock Microsoft {COCO:} Common Objects in Context.
\newblock In \emph{Proceedings of the ECCV}, 740--755.

\bibitem[{Liu et~al.(2021)Liu, Lin, Cao, Hu, Wei, Zhang, Lin, and
  Guo}]{liu2021Swin}
Liu, Z.; Lin, Y.; Cao, Y.; Hu, H.; Wei, Y.; Zhang, Z.; Lin, S.; and Guo, B.
  2021.
\newblock Swin Transformer: Hierarchical Vision Transformer using Shifted
  Windows.
\newblock \emph{arXiv preprint arXiv:2103.14030}.

\bibitem[{Lu et~al.(2019)Lu, Batra, Parikh, and Lee}]{DBLP:conf/nips/LuBPL19}
Lu, J.; Batra, D.; Parikh, D.; and Lee, S. 2019.
\newblock ViLBERT: Pretraining Task-Agnostic Visiolinguistic Representations
  for Vision-and-Language Tasks.
\newblock In \emph{Proceedings of the NeurIPS}, 13--23.

\bibitem[{Luo et~al.(2021)Luo, Ji, Sun, Cao, Wu, Huang, Lin, and
  Ji}]{DBLP:conf/aaai/LuoJSCWHLJ21}
Luo, Y.; Ji, J.; Sun, X.; Cao, L.; Wu, Y.; Huang, F.; Lin, C.; and Ji, R. 2021.
\newblock Dual-level Collaborative Transformer for Image Captioning.
\newblock In \emph{Proceedings of the AAAI}, 2286--2293. {AAAI} Press.

\bibitem[{Pan et~al.(2020)Pan, Yao, Li, and Mei}]{DBLP:conf/cvpr/PanYLM20}
Pan, Y.; Yao, T.; Li, Y.; and Mei, T. 2020.
\newblock X-Linear Attention Networks for Image Captioning.
\newblock In \emph{Proceedings of the CVPR}, 10968--10977.

\bibitem[{Papineni et~al.(2002)Papineni, Roukos, Ward, and
  Zhu}]{DBLP:conf/acl/PapineniRWZ02}
Papineni, K.; Roukos, S.; Ward, T.; and Zhu, W. 2002.
\newblock Bleu: a Method for Automatic Evaluation of Machine Translation.
\newblock In \emph{Proceedings of the ACL}, 311--318.

\bibitem[{Radford et~al.(2021)Radford, Kim, Hallacy, Ramesh, Goh, Agarwal,
  Sastry, Askell, Mishkin, Clark, Krueger, and
  Sutskever}]{DBLP:conf/icml/RadfordKHRGASAM21}
Radford, A.; Kim, J.~W.; Hallacy, C.; Ramesh, A.; Goh, G.; Agarwal, S.; Sastry,
  G.; Askell, A.; Mishkin, P.; Clark, J.; Krueger, G.; and Sutskever, I. 2021.
\newblock Learning Transferable Visual Models From Natural Language
  Supervision.
\newblock In Meila, M.; and Zhang, T., eds., \emph{Proceedings of the ICML},
  volume 139 of \emph{Proceedings of Machine Learning Research}, 8748--8763.
  {PMLR}.

\bibitem[{Ren et~al.(2017)Ren, He, Girshick, and Sun}]{DBLP:conf/nips/RenHGS15}
Ren, S.; He, K.; Girshick, R.~B.; and Sun, J. 2017.
\newblock Faster {R-CNN:} Towards Real-Time Object Detection with Region
  Proposal Networks.
\newblock \emph{{IEEE} Trans. Pattern Anal. Mach. Intell.}, 39(6): 1137--1149.

\bibitem[{Rennie et~al.(2017)Rennie, Marcheret, Mroueh, Ross, and
  Goel}]{DBLP:conf/cvpr/RennieMMRG17}
Rennie, S.~J.; Marcheret, E.; Mroueh, Y.; Ross, J.; and Goel, V. 2017.
\newblock Self-Critical Sequence Training for Image Captioning.
\newblock In \emph{Proceedings of the CVPR}, 1179--1195.

\bibitem[{Simonyan and Zisserman(2015)}]{DBLP:journals/corr/SimonyanZ14a}
Simonyan, K.; and Zisserman, A. 2015.
\newblock Very Deep Convolutional Networks for Large-Scale Image Recognition.
\newblock In Bengio, Y.; and LeCun, Y., eds., \emph{Proceedings of the ICLR}.

\bibitem[{Vaswani et~al.(2017)Vaswani, Shazeer, Parmar, Uszkoreit, Jones,
  Gomez, Kaiser, and Polosukhin}]{DBLP:conf/nips/VaswaniSPUJGKP17}
Vaswani, A.; Shazeer, N.; Parmar, N.; Uszkoreit, J.; Jones, L.; Gomez, A.~N.;
  Kaiser, L.; and Polosukhin, I. 2017.
\newblock Attention is All you Need.
\newblock In \emph{Proceedings of the NIPS}, 5998--6008.

\bibitem[{Vedantam, Zitnick, and Parikh(2015)}]{DBLP:conf/cvpr/VedantamZP15}
Vedantam, R.; Zitnick, C.~L.; and Parikh, D. 2015.
\newblock CIDEr: Consensus-based image description evaluation.
\newblock In \emph{Proceedings of the CVPR}, 4566--4575.

\bibitem[{Vinyals et~al.(2015)Vinyals, Toshev, Bengio, and
  Erhan}]{DBLP:conf/cvpr/VinyalsTBE15}
Vinyals, O.; Toshev, A.; Bengio, S.; and Erhan, D. 2015.
\newblock Show and tell: {A} neural image caption generator.
\newblock In \emph{Proceedings of the CVPR}, 3156--3164.

\bibitem[{Wang, Chen, and Hu(2019)}]{DBLP:conf/aaai/WangC019}
Wang, W.; Chen, Z.; and Hu, H. 2019.
\newblock Hierarchical Attention Network for Image Captioning.
\newblock In \emph{Proceedings of the AAAI}, 8957--8964.

\bibitem[{Xu et~al.(2015)Xu, Ba, Kiros, Cho, Courville, Salakhutdinov, Zemel,
  and Bengio}]{DBLP:conf/icml/XuBKCCSZB15}
Xu, K.; Ba, J.; Kiros, R.; Cho, K.; Courville, A.~C.; Salakhutdinov, R.; Zemel,
  R.~S.; and Bengio, Y. 2015.
\newblock Show, Attend and Tell: Neural Image Caption Generation with Visual
  Attention.
\newblock In \emph{Proceedings of the ICML}, 2048--2057.

\bibitem[{Yao et~al.(2018)Yao, Pan, Li, and Mei}]{DBLP:conf/eccv/YaoPLM18}
Yao, T.; Pan, Y.; Li, Y.; and Mei, T. 2018.
\newblock Exploring Visual Relationship for Image Captioning.
\newblock In \emph{Proceedings of the ECCV}, 711--727.

\bibitem[{Zhang et~al.(2021)Zhang, Sun, Luo, Ji, Zhou, Wu, Huang, and
  Ji}]{DBLP:conf/cvpr/ZhangSLJZWHJ21}
Zhang, X.; Sun, X.; Luo, Y.; Ji, J.; Zhou, Y.; Wu, Y.; Huang, F.; and Ji, R.
  2021.
\newblock RSTNet: Captioning With Adaptive Attention on Visual and Non-Visual
  Words.
\newblock In \emph{Proceedings of the CVPR}, 15465--15474. Computer Vision
  Foundation / {IEEE}.

\bibitem[{Zhu and Yang(2020)}]{DBLP:conf/cvpr/ZhuY20a}
Zhu, L.; and Yang, Y. 2020.
\newblock ActBERT: Learning Global-Local Video-Text Representations.
\newblock In \emph{Proceedings of the CVPR}, 8743--8752. {IEEE}.

\end{thebibliography}

\end{document}